\documentclass[10pt,twocolumn,letterpaper]{article}
\usepackage{iccv}
\usepackage{times}
\usepackage{epsfig}
\usepackage{graphicx}
\usepackage{amsmath}
\usepackage{amssymb}
\usepackage{gensymb}

\usepackage[breaklinks=true,bookmarks=false]{hyperref}

\iccvfinalcopy 

\def\iccvPaperID{8} 

\ificcvfinal\fi
\begin{document}

\title{Where to Look Next: Unsupervised Active Visual Exploration on 360\degree  Input}

\author{Soroush Seifi\hspace{0,5cm}Tinne Tuytelaars\\
KU Leuven, ESAT-PSI\\
Kasteelpark Arenberg 10, 3001 Leuven, Belgium\\
{\tt\small \{sseifi, tinne.tuytelaars\}@esat.kuleuven.be}
}

\clearpage\maketitle
\thispagestyle{empty}

\begin{abstract}
\vspace{-0,3cm}
 We address the problem of active visual exploration of large 360\degree inputs. In our setting an active agent with a limited camera bandwidth explores its 360\degree environment by changing its viewing direction at limited discrete time steps. As such, it observes the world as a sequence of narrow field-of-view `glimpses', deciding for itself where to look next.
Our proposed method exceeds previous works' performance by a significant margin without the need for deep reinforcement learning or training separate networks as sidekicks. 
A key component of our system are the spatial memory maps that make the system aware of the glimpses' orientations (locations in the 360\degree image).
Further, we stress the advantages of retina-like glimpses when the agent's sensor-bandwidth and time-steps are limited.
Finally, we use our trained model to do classification of the whole scene using only the information observed in the glimpses.
\end{abstract}
\vspace{-0,6cm}
\section{Introduction}
\vspace{-0,25cm}

In this paper, we address the active exploration problem defined in \cite{c1,c2} where an agent with a limited camera sensor bandwidth  and limited field of view changes its viewing direction sequentially at discrete time-steps to acquire a maximum amount of information from its 360\degree environment. At each time-step, the agent sees a cropped part (called a 'glimpse') of the panorama representing its environment.
Therefore, in order to understand its environment, the agent needs to combine information from multiple glimpses, correlate the spatial locations of the glimpses it has seen, fill in the missing parts and decide where to look next.

This relatively new problem setting can build on results obtained in
the visual attention literature, where often
an encoder-decoder architecture is used with Recurrent Neural Networks (e.g.~LSTMs) to compress the extracted features from attended regions \cite{c3,c20,c21,c22}.
However an additional challenge in the active exploration setting defined above is inferring the spatial correlation of glimpses in 2D. In \cite{c1,c2}, the relative location of glimpses is fed to the system using 
image coordinates.
We argue that it is difficult for the LSTM cells to correlate the visual information using explicit relative coordinates. This results in loss of texture and details in the reconstructions of the input panorama in previous works. Getting inspiration from \cite{c36,c37} where 2D memory architectures are shown to be effective for navigation in partially observable environments, we replace the LSTM cells in \cite{c1,c2} with our proposed {\em spatial memory maps}. Such memory maps are simpler in architecture compared to those in \cite{c36,c37} and maintain all extracted information from glimpses for all time-steps in their correct spatial location. This will guarantee that the spatial correlations of the glimpses are kept intact. 

\begin{figure}[t]
\begin{center}
\includegraphics[width=\linewidth]{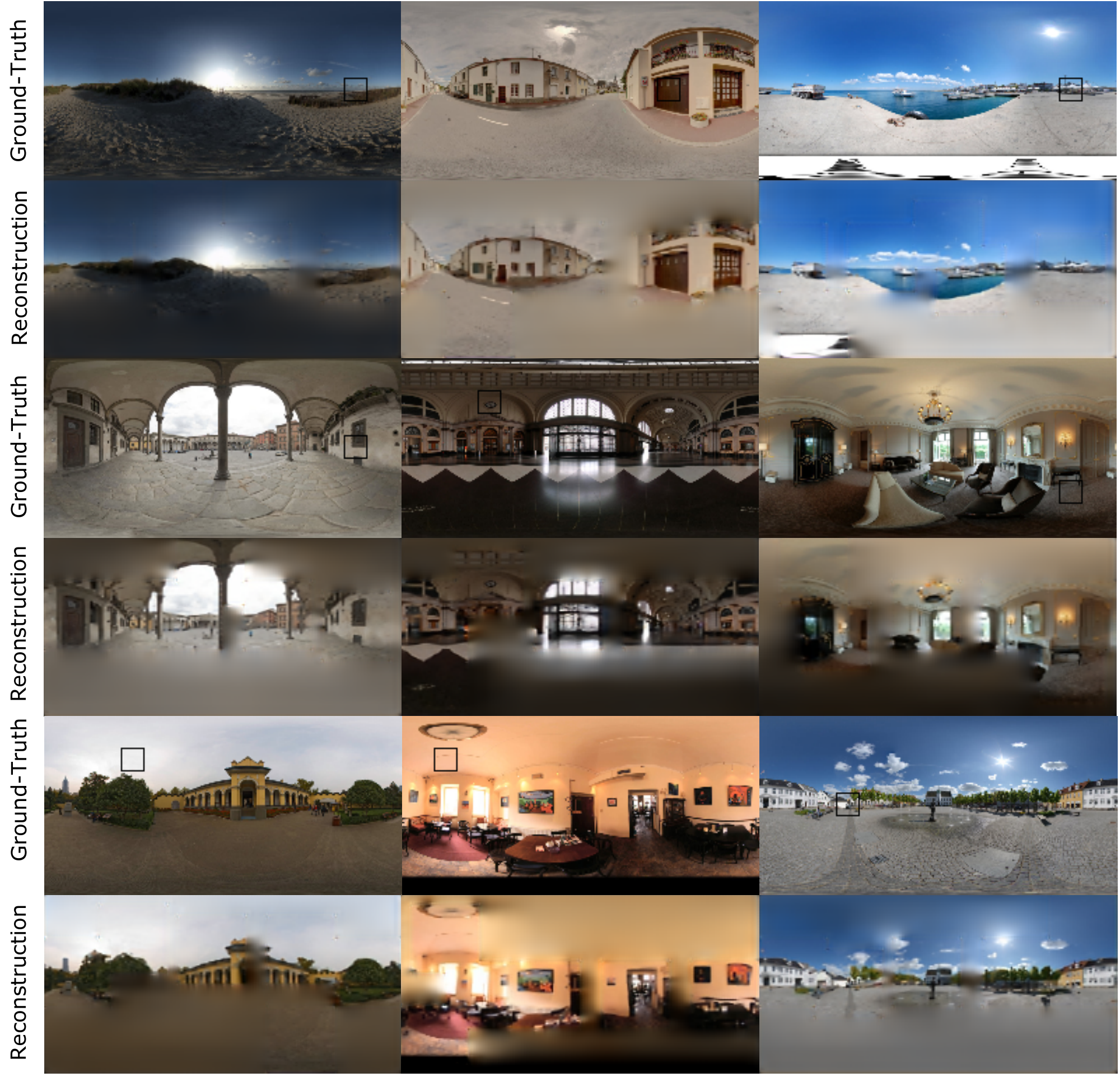}
\end{center}
\vspace{-0,3cm}
   \caption{Our model's reconstructions (rows 2,4,6) in comparison to the ground-truth (rows 1,3,5), after a sequence of 6 glimpses covering 14\% of image pixels.
   Zoom-in for best observation.
   } 
   \vspace{-0,6cm}
\label{reconst-truth}
\end{figure}

Furthermore, \cite{c1,c2} use deep reinforcement learning to train a policy to produce the sequence of glimpses. \cite{c2} trains separate networks using the full input image to help learning such policy during training. Instead, our model is directly trained with the reconstruction loss, avoiding the hard-to-train reinforcement learning and the need for observability imbalance between training  and test time. 

Our architecture can be particularly useful for applications where transmitting the whole input image is not possible due to bandwidth limitations. In this case, one could use our architecture to select the areas that make the reconstruction at the destination easier.
%

\begin{figure*}
\vspace{-0,55cm}
\begin{center}
\includegraphics[width=\linewidth]{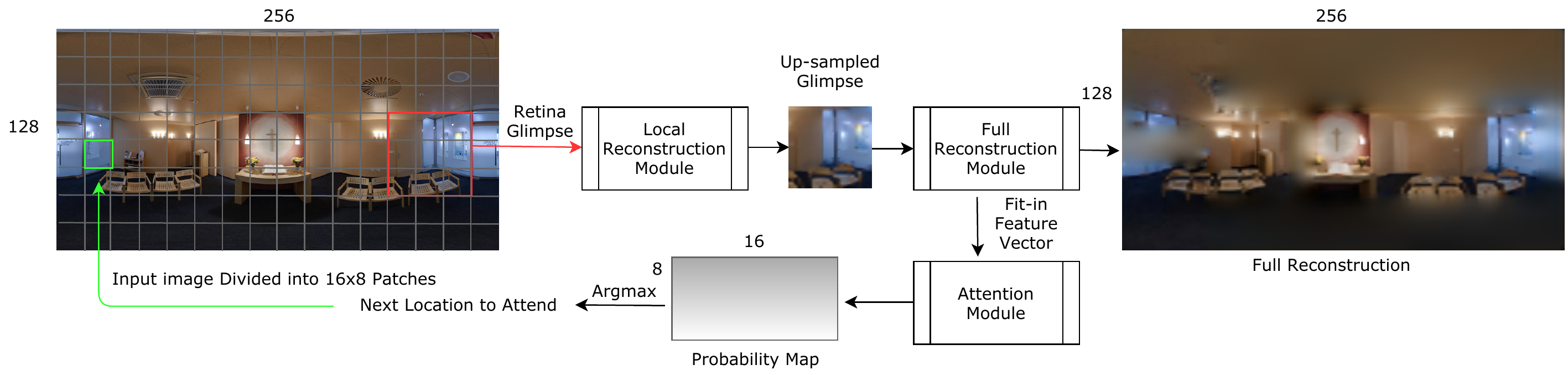}
\end{center}
\vspace{-0,35cm}
   \caption{Overview of our architecture at an arbitrary time-step.
   } 
   \vspace{-0,35cm}
\label{architecture-overview}
\end{figure*}

Figure \ref{reconst-truth} shows some of our results on the test data from the SUN360 dataset \cite{c38}. Figure \ref{architecture-overview} illustrates the overall architecture of our network. Since the focus of this paper is not generating photo realistic reconstructions, we did not use any sophisticated method of in-painting for reconstruction. However, combining our architecture with an in-painting method such as \cite{c28,c29,c33} can generate even better reconstructions of the input panorama.

The remainder of the paper is organized as follows. In section 2, we define our method. We describe our experimental results in section 3. Section 4 concludes the paper.

\vspace{-0,3cm}
\section{Method}
\vspace{-0,2cm}
Our architecture consists of three main modules
(see also Figure~\ref{architecture-overview}) described below:
 
\noindent \textbf{- Local Reconstruction Module:}
We exploit the camera bandwidth more efficiently with the help of retina-like glimpses, as used in the visual attention literature~\cite{c3,c4}. In this case, the central part of each glimpse is sampled in full resolution while the remaining pixels on each side of the glimpse are downsampled by a factor 2.

\begin{figure}
\begin{center}
\includegraphics[width=\linewidth]{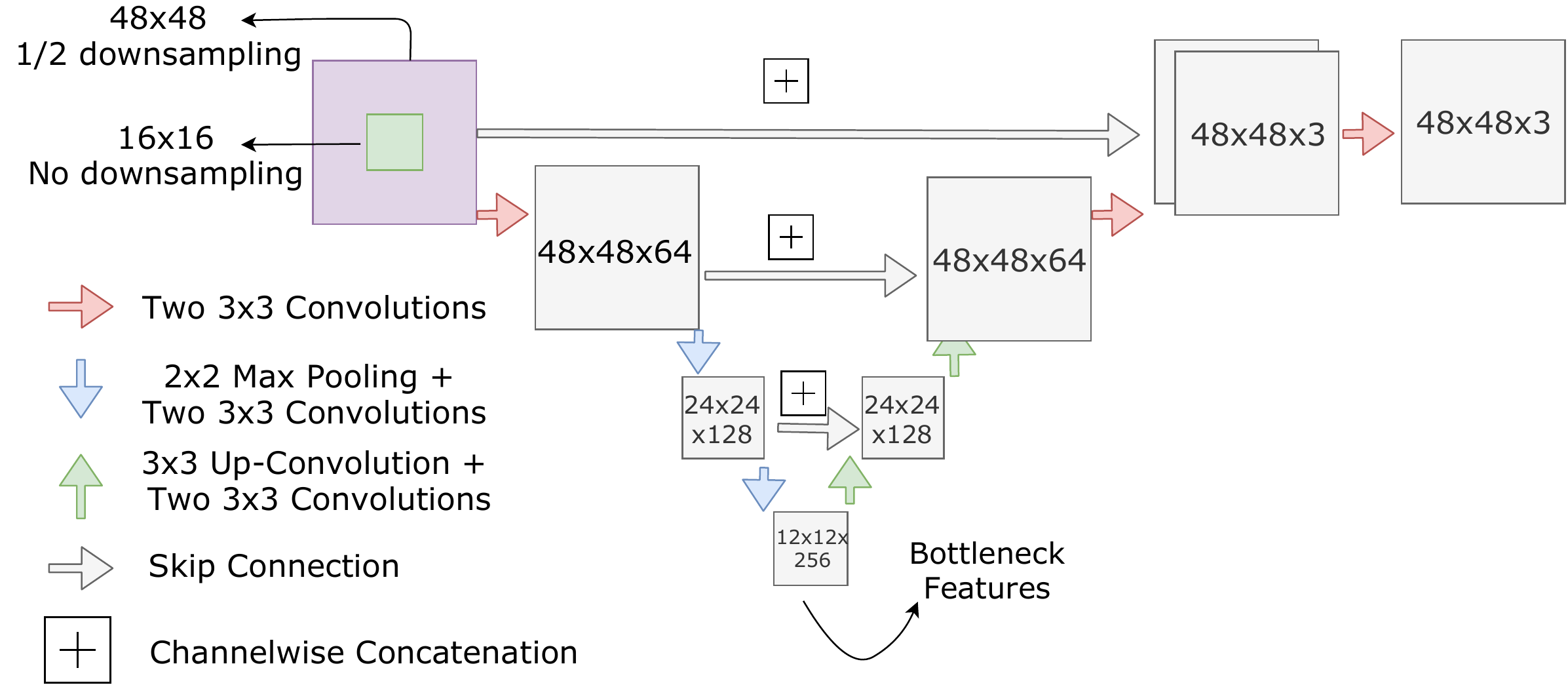}
\end{center}
\vspace{-0,5cm}
   \caption{Local reconstruction module's architecture}
\label{unet-arch}
\end{figure}

We modify a super-resolution architecture known as U-net~\cite{c5} to bring these partly down-sampled glimpses to their full resolution in the 'local reconstruction module', details of which are found in Figure \ref{unet-arch}.

We use L1 loss between the ground-truth glimpse and reconstructed glimpse at each time-step. 
%
The up-sampled glimpses and their corresponding bottleneck features 
are next passed to the full reconstruction module.

\noindent \textbf{- Full Reconstruction Module:}
This module fits each up-sampled glimpse in a {\em spatial memory map} denoted as {\em fit-in matrix}.
It has the same dimensions as the input panorama and all its values are initially set to 0. At each time-step, the area corresponding to the glimpse location gets filled with the up-sampled glimpse. 

Additionally, we further sample down the bottleneck features for each glimpse to $4\times4\times8$ using average pooling and convolutions. We flatten these features and fit them in a {\em fit-in feature vector}. This partially filled vector of size 4096 keeps our encoded representation for the whole input panorama according to the visited glimpses.
We use fully-connected layers on top of this vector to reconstruct a down-sampled version of the input panorama at each step. We denote this reconstruction as our {\em background reconstruction}. 

\begin{figure}
  \begin{center}
  \vspace{-0,25cm}
      \includegraphics[width=\linewidth]{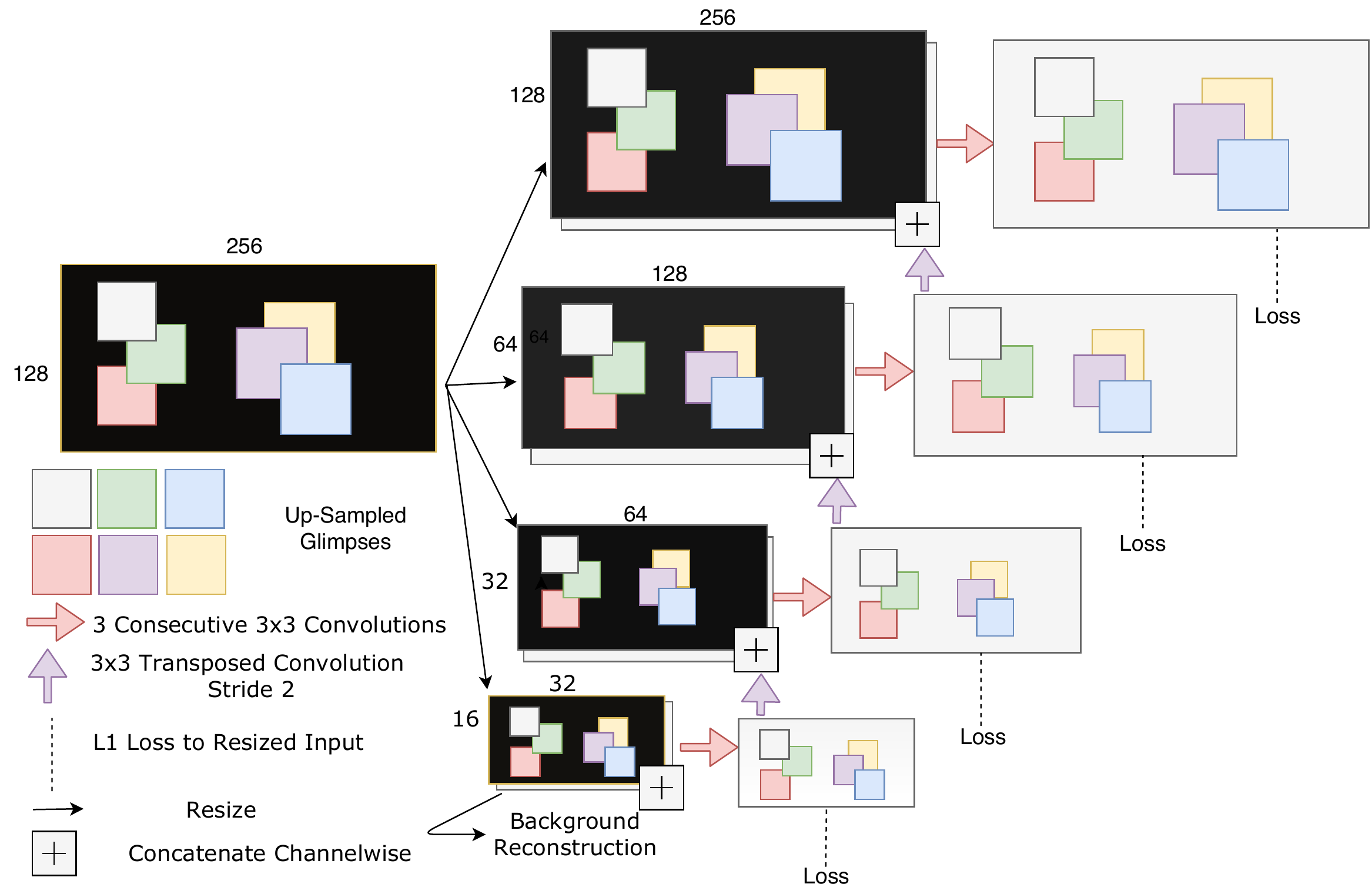}
   \end{center}
   \vspace{-0,3cm}
      \caption{Input reconstruction using background reconstruction and skip-connections.}
      \vspace{-0,35cm}
      \label{up-sample-fullres}
   \end{figure}

The background reconstruction is then scaled up to the input size with transposed convolutions and using the fit-in matrix as skip connection in different scales (see Figure~\ref{up-sample-fullres}). This way, the network learns to paste the reconstructed glimpses from the fit-in matrix while relying on the background reconstruction for in-painting the unseen areas. 
We optimize L1 loss between the input panorama and the reconstruction for each scale to train this module.


\noindent \textbf{- Attention Module:}
Finally, the attention module determines where to look next.
The reconstruction loss at each time-step can inform the model about the areas in the input where the model's representation encodes the least information in the panorama. Therefore, the area with the highest reconstruction loss can be selected for attendance in the next time-step. Our attention module is trained to predict such area at test time. In particular, we divide the image to non-overlapping $16\times16$ patches (128 image patches in total). We use fully connected layers on top of our fit-in feature vector to predict the probability for each patch to have the highest accumulative reconstruction error. During training, we determine the ground-truth probability for each image patch by summing up the reconstruction loss for all its pixels and normalizing it by the accumulative loss for all pixels in the panorama. We use a sparse softmax cross-entropy loss
for optimizing the attention module.

In order to stimulate exploration during training time, we use a multinomial distribution to sample the next location to attend from our predicted probability map. However, at test time we always take the location with highest probability in the probability map.

\begin{figure*}
\begin{center}
\includegraphics[width=\linewidth]{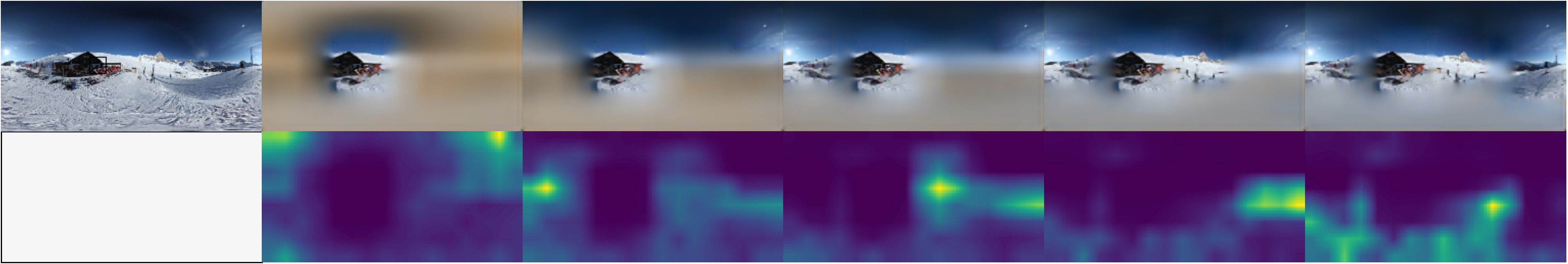}
\end{center}
\vspace{-0,3cm}
   \caption{Evolution of the reconstruction for 5 time-steps. First row: Full-reconstruction at each time step. Second row: Attention probability map for the next location; the brighter a pixel, the higher the probability for attendance. First column: Ground-truth image and uniform probability for all locations (first glimpse is selected randomly).}  
   \vspace{-0,4cm}
\label{step-by-step}
\end{figure*}


\noindent \textbf{Transfer Learning:}
Apart from the reconstruction task, our method can be extended to other tasks with dense per-pixel loss available. Furthermore, since the reconstruction error encodes the model's uncertainty about image regions, by setting reconstruction as an auxiliary task and using reconstruction error to train the attention module, the architecture can still attend meaningful regions regardless of the number of labels for the main task.

To evaluate our method in such scenario, we modify our architecture for predicting the scene category for each image after taking a fixed number of glimpses. In one setting we feed the final reconstruction to a classification network (VGG-19 \cite{c17}) and train the whole architecture 
end-to-end 
to predict the class labels. In another setting, the network attends the locations suggested by the attention module and fits extracted features to a {\em fit-in classification vector}. We use fully-connected layers followed by a softmax function on top of this feature vector after visiting the last glimpse to predict the class labels.
This way we separate the classification network from the reconstruction network for optimization. Besides, we train a VGG-19 network on the whole input image as an upper-bound.



\noindent \textbf{Optimization:}
We train our network end-to-end with Adam optimizer \cite{c34} using the loss function:
\begin{equation}
    Loss=\sum_{t=1}^{T}L_{local}^{t}+\sum_{(m,n)\in S}L_{m\times n}+L_{attention}
    \label{full-loss}
\end{equation}
where S = \{(16,32),(32,64),(64,128),(128,256)\} represents the different scales for the full reconstruction module and $L_{local}^{t}$, $L_{m\times n}$ and $L_{attention}$ correspond to the losses for the local reconstruction, full reconstruction and attention module respectively.
We obtained best results by optimizing the local reconstruction module for every time-step while the attention and full reconstruction modules are optimized only for the last time-step.

For classification, 
a cross-entropy classification loss is added to the right-hand side of equation \ref{full-loss}.
\vspace{-0,2cm}
\section{Experiments}
\vspace{-0,1cm}
We evaluate our method on 26 categories of SUN360 dataset used in \cite{c1,c2}. Since SUN360 dataset consists of images with continuity along the x-axis, we augment the data and generate new training examples by cropping a part from one side of the image and pasting it to the other side.

In our experiments, we divide each image in a $16\times8$ grid of $16\times16$ blocks (see also Figure~\ref{architecture-overview}).
At each time-step we take a retina-like glimpse centered on a block from this grid. The central $16\times16$ part of each glimpse is sampled in full resolution. However, 
the 8 neighbouring blocks are downsampled by a factor of 2
(see also Figure~\ref{unet-arch}). This way by taking 8 glimpses per image we remain consistent in terms of amount of information with previous works which took 6 glimpses of size $32\times32$ in full resolution. 



We report our model's performance using Root Mean Squared Error (RMSE) in range (0-255). 
For consistency with previous works~\cite{c1,c2}, we also 
report the Mean Squared Error (MSE) in range (0-1)$\times$1000 -- see table \ref{comparison-table}. 
As a reference, we also provide results for baselines where glimpses locations are chosen randomly either from all possible locations or only from the middle row in the image. We also report results for a baseline where the next glimpse is chosen from a block neighbouring the current glimpse (same setting as \cite{c1,c2}). Our method outperforms all previous works and baselines and its performance is close to an upper-bound where glimpses are attended based on ground-truth reconstruction error rather than relying on our attention module's predictions.

Figure \ref{step-by-step} illustrates our model's outputs for 5 time-steps. Our model gains a general understanding of the environment in the first few glimpses (e.g sky/ground layout). Afterwards, it starts scanning the horizon which is predicted to have more details.

\begin{figure}
\begin{center}
    \includegraphics[width=1.05\linewidth]{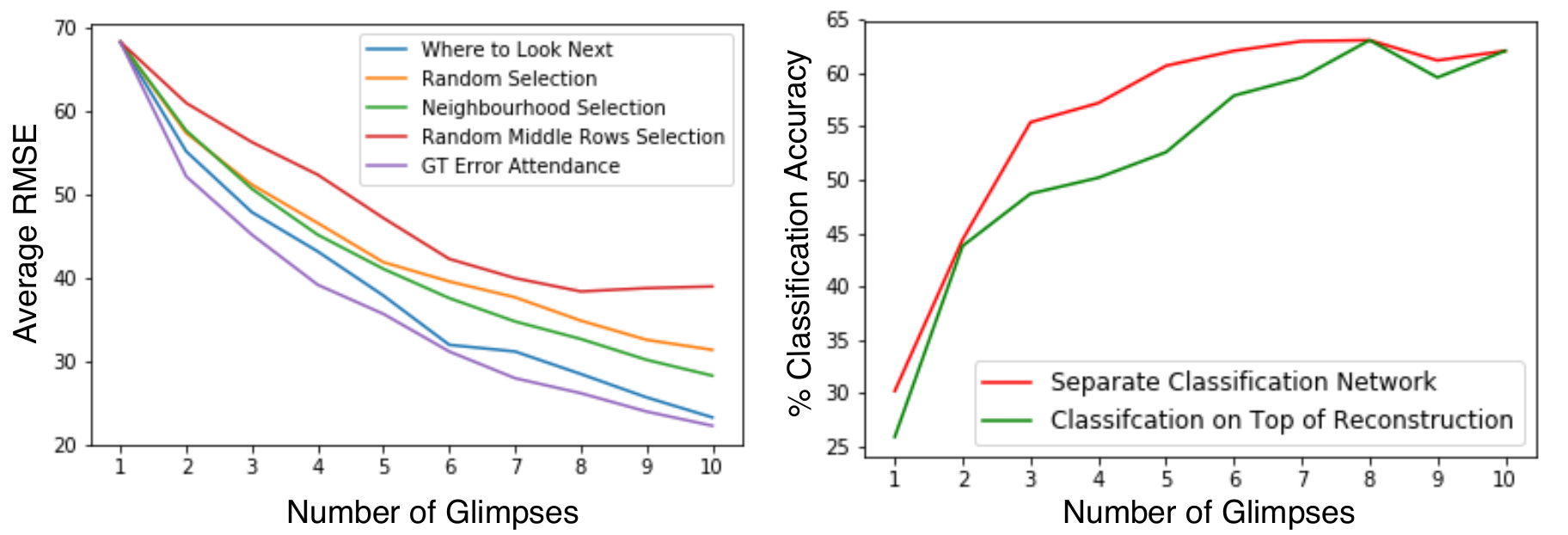}
\end{center}
\vspace{-0,3cm}
  \caption{Left: Where to look next's performance compared to the baselines. Right: Classification accuracy with transfer learning.}
\vspace{-0,5cm}
  \label{mean-loss-steps}
\end{figure}

Figure \ref{mean-loss-steps} (Left) illustrates the improvement in reconstruction for different time-steps. With more glimpses the performance improves and the gap between our approach and the baselines gets bigger. We exceed the previous works' performance by taking even 5 glimpses which covers only 11\% of pixels in the image (19\% in previous works). Besides, in previous works, each glimpse is initially projected to a normal field of view. \cite{c35} shows that such a glimpse covers a larger area of the input panorama compared to our setting where we crop the glimpse from the panorama without any projection step. Therefore, our model outperforms previous works with even harder constraints.


Training a network for classification based on the whole input images results in 65\% accuracy. Figure~\ref{mean-loss-steps} (Right) shows that with 8 glimpses, we 
get close to that result, while using only a fraction of the image content.


\begin{table}
\begin{center}
{\footnotesize 
\begin{tabular}{|l|c|c|}
\hline
Method&MSE&RMSE\\
&[0-1]$\times$1000& [0-255]\\
\hline\hline
Side-kick Policy Learning \cite{c1} &23.36 & 39.0  \\ 
Learning to Look Around \cite{c2} & 23.16 & 38.8 \\
Where to Look Next (ours)& {\bf 12.49}  & {\bf 28.5}\\ 
\hspace{0,3cm} with Random Selection & 18.73 & 34.9\\
\hspace{0,3cm} with Middle Rows Random Selection  & 22.67 & 46.7  \\
\hspace{0,3cm} with Neighbourhood Selection & 16.64 & 32.9 \\ 
\hspace{0,3cm} with GT Error Attendance (upper bound) & 10.39 & 26.0\\ 
\hline
\end{tabular}
}
\end{center}
\vspace{-0,2cm}
\caption{Reconstruction error for different methods and baselines on the SUN360 dataset, evaluated using RMSE in [0-255] 
and with the metric used in~\cite{c1,c2} 
.}
\vspace{-0,5cm}
\label{comparison-table}
\end{table}

\vspace{-0,2cm}
\section{Conclusion}
\vspace{-0,2cm}
In this paper we defined a new architecture for active exploration of an agent in its 360\degree environment. We addressed some of the limitations from previous works by effectively using the agent's camera bandwidth, replacing LSTM cells with spatial memory maps and optimizing our network directly with the reconstruction loss. This way, we outperformed previous work's performance by a margin. Finally, we further evaluated our method by solving a classification task reaching results close to an upper bound.
\vspace{-0,45cm}
\paragraph{Acknowledgment}
This work was supported by the FWO SBO project Omnidrone \footnote{https://www.omnidrone720.com/}.

\vspace{-0,3cm}


\begin{thebibliography}{99}
\vspace{-0,25cm}
\bibitem{c1} Jayaraman D., Grauman K. Learning to look around: intelligently exploring unseen environments for unknown tasks. 
CVPR 2018 (pp. 1238-1247).
\vspace{-0,25cm}
\bibitem{c2} Ramakrishnan S.K., Grauman K. Sidekick policy learning for active visual exploration.ECCV 2018 (pp. 413-430).
\vspace{-0,2cm}
\bibitem{c3} Xu K., Ba J., Kiros R., Cho K., Courville A., Salakhudinov R., Zemel R., Bengio Y. Show, attend and tell: Neural image caption generation with visual attention.
 ICML 2015
 (pp. 2048-2057).
\vspace{-0,25cm}
\bibitem{c20}Ba J., Mnih V., Kavukcuoglu K. Multiple object recognition with visual attention. ICLR 2015.
\vspace{-0,25cm}
\bibitem{c21}Gregor K., Danihelka I., Graves A., Rezende D.J., Wierstra D. Draw: A recurrent neural network for image generation. ICML 2015.
\vspace{-0,25cm}
\bibitem{c22}Mnih V., Heess N., Graves A. Recurrent models of visual attention. NIPS 2014 (pp. 2204-2212).
\vspace{-0,25cm}
\bibitem{c36} Parisotto E, Salakhutdinov R. Neural map: Structured memory for deep reinforcement learning. ICLR 2017.

\bibitem{c37} Henriques JF, Vedaldi A. Mapnet: An allocentric spatial memory for mapping environments. Inproceedings of CVPR 2018 (pp. 8476-8484).
\vspace{-0,25cm}
\bibitem{c38}Xiao J, Ehinger KA, Oliva A, Torralba A. Recognizing scene viewpoint using panoramic place representation. In CVPR 2012 (pp. 2695-2702)
\vspace{-0,25cm}
\bibitem{c28}Liu G., Reda F.A., Shih K.J., Wang T.C., Tao A., Catanzaro B. Image inpainting for irregular holes using partial convolutions.
ECCV 2018 (pp. 85-100).
\vspace{-0,25cm}
\bibitem{c29} Oord A.V., Kalchbrenner N., Kavukcuoglu K. Pixel recurrent neural networks. ICML 2016.
\vspace{-0,25cm}
\bibitem{c33}Yu J., Lin Z., Yang J., Shen X., Lu X., Huang T.S. Generative image inpainting with contextual attention. CVPR 2018 (pp. 5505-5514).
\vspace{-0,25cm}
\bibitem{c4} Mnih V., Heess N., Graves A. Recurrent models of visual attention. 
NIPS 2014 (pp. 2204-2212).
\vspace{-0,25cm}
\bibitem{c5} Ronneberger O, Fischer P, Brox T. U-net: Convolutional networks for biomedical image segmentation. MICCAI 2015 (pp. 234-241). 
\vspace{-0,25cm}
\bibitem{c17} Simonyan K., Zisserman A. Very deep convolutional networks for large-scale image recognition. ICLR 2015. 
\vspace{-0,25cm}
\bibitem{c34}Kingma D.P., Ba J. Adam: A method for stochastic optimization. ICLR 2015.
\vspace{-0,25cm}
\bibitem{c35}
Ramakrishnan SK, Jayaraman D, Grauman K. Emergence of exploratory look-around behaviors through active observation completion. Science Robotics. 2019 May 15;4(30):eaaw6326.
 \vspace{-0,25cm}
\end{thebibliography}
\end{document}